\documentclass{article} 
\usepackage{collas2024_conference,times}
\usepackage{easyReview}
\usepackage{algorithm}
\usepackage{algpseudocode}
\usepackage[acronym, hyperfirst=false]{glossaries}
\usepackage{hyperref}
\hypersetup{
    colorlinks=true,
    linkcolor=red,
    filecolor=magenta,
    urlcolor=blue,
    citecolor=purple,
    pdftitle={Overleaf Example},
    pdfpagemode=FullScreen,
    }
\usepackage{url}
\usepackage{tikz}
\usepackage[utf8]{inputenc} 
\usepackage[T1]{fontenc}    
\usepackage{url}            
\usepackage{booktabs}       
\usepackage{amsfonts}       
\usepackage{nicefrac}       
\usepackage{microtype}      
\usepackage{multirow}
\usepackage{graphicx}
\usepackage{doi}
\usepackage{amsmath,amssymb,amsthm} 
\usepackage{enumitem} 
\usepackage{wrapfig}
\usepackage{float}
\usepackage{pifont}
\usepackage{multirow}
\usepackage{euscript}
\usepackage[many]{tcolorbox}
\usepackage{datetime}
\usepackage{placeins}
\usepackage{tikz}
\usepackage{stmaryrd}
\usepackage{mathtools}
\usepackage{tabularx}
\usepackage{bbm}
\usepackage{cleveref}
\usepackage{caption}
\usepackage{mdframed}
\usepackage[fixed]{fontawesome5}

\title{Infinite dSprites for Disentangled Continual Learning: Separating Memory Edits from Generalization}


\author{
\makebox[\textwidth][c]{
\textbf{Sebastian Dziadzio}$^{1, 2}$\quad\quad \textbf{\c{C}a\u{g}atay Y{\i}ld{\i}z}$^{1, 2}$\quad\quad \textbf{Gido M. van de Ven}$^{3}$
} \\ \\
\makebox[\textwidth][c]{
\textbf{Tomasz Trzciński}$^{4, 5, 6}$\quad\quad \textbf{Tinne Tuytelaars}$^{3}$\quad\quad \textbf{Matthias Bethge}$^{1, 2}$
} \\ \\
{\small $^1$T\"ubingen AI Center $^2$University of T\"ubingen $^3$KU Leuven $^4$IDEAS NCBR $^5$Tooploox $^6$Warsaw University of Technology}
}

\collasfinalcopy

\setacronymstyle{long-short}
\newacronym{ewc}{EWC}{Elastic Weight Consolidation}
\newacronym{si}{SI}{Synaptic Intelligence}
\newacronym{lwf}{LwF}{Learning without Forgetting}
\newacronym{vcl}{VCL}{Variational Continual Learning}
\newacronym{agem}{A-GEM}{Averaged Gradient Episodic Memory}
\newacronym{gem}{GEM}{Gradient Episodic Memory}
\newacronym{der}{DER}{Dark Experience Replay}
\newacronym{l2p}{L2P}{Learning to Prompt}
\newacronym{ids}{idSprites}{Infinite dSprites}
\newacronym{dcl}{DCL}{Disentangled Continual Learning}
\newacronym{lda}{LDA} {Linear Discriminate Analysis}
\newglossaryentry{fov}
{
  name={FoV},
  description={factors of variation},
  first={factor of variation (FoV)},
  plural={FoVs},
  firstplural={factors of variation (FoVs)}
}

\begin{document}

\maketitle

\vspace{-0.5cm}
\begin{center}
    \raisebox{-1pt}{\faGithub} \href{https://github.com/sbdzdz/disco}{\fontsize{8.8pt}{0pt}\path{github.com/sbdzdz/disco}} \quad {\faGithub}\href{https://github.com/sbdzdz/idsprites}{\fontsize{8.8pt}{0pt} \path{github.com/sbdzdz/idsprites}} \\
\end{center}
\vspace{0.7cm}

\begin{abstract}
 The ability of machine learning systems to learn continually is hindered by catastrophic forgetting, the tendency of neural networks to overwrite previously acquired knowledge when learning a new task. Existing methods mitigate this problem through regularization, parameter isolation, or rehearsal, but they are typically evaluated on benchmarks comprising only a handful of tasks. In contrast, humans are able to learn over long time horizons in dynamic, open-world environments, effortlessly memorizing unfamiliar objects and reliably recognizing them under various transformations. To make progress towards closing this gap, we introduce Infinite dSprites, a parsimonious tool for creating continual classification and disentanglement benchmarks of arbitrary length and with full control over generative factors. We show that over a sufficiently long time horizon, the performance of all major types of continual learning methods deteriorates on this simple benchmark. This result highlights an important and previously overlooked aspect of continual learning: given a finite modelling capacity and an arbitrarily long learning horizon, efficient learning requires memorizing class-specific information and accumulating knowledge about general mechanisms. In a simple setting with direct supervision on the generative factors, we show how learning class-agnostic transformations offers a way to circumvent catastrophic forgetting and improve classification accuracy over time. Our approach sets the stage for continual learning over hundreds of tasks with explicit control over memorization and forgetting, emphasizing open-set classification and one-shot generalization.
\end{abstract}

\section{Introduction}
\label{sec:intro}

Continual learning methods are typically evaluated using standard computer vision datasets segmented into disjoint classification tasks. Despite advancements in model scale and complexity, even recent continual learning benchmarks like Split-ImageNet-R \citep{wang2022dualprompt} feature a limited number of tasks and classes. This constraint raises concerns about their ability to accurately reflect open-ended learning scenarios that humans routinely tackle. To address this issue, we introduce \acrfull{ids}, a continual learning benchmark generator inspired by the dSprites dataset \citep{dsprites17}. It procedurally generates a virtually infinite progression of two-dimensional shapes in every combination of orientation, scale, and position~(see \cref{fig:idsprites}). Crucially, by providing ground truth values of individual \glspl{fov}, \acrshort{ids} paves the way for methods that efficiently exploit a fundamental property of many real-world continual classification problems, namely the compositional interactions between objects and transformations.  In the spirit of the Omniglot Challenge \citep{lake2015human}, and the Abstraction and Reasoning Corpus \citep{chollet2019measure}, we envision solving \acrshort{ids} as a stepping stone towards continual learning systems that demonstrate human-like intelligence, characterized by sample efficiency, compositional representations, and the ability to perform one-shot generalization and lifelong open-set learning.

We show that all major types of continual learning methods eventually break down on this simple benchmark. This finding indicates that strategies such as regularization, parameter isolation, experience replay, or parameter-efficient adaptation are not sufficient to mitigate catastrophic forgetting in a truly lifelong learning setting. We further demonstrate that even state of the art vision-language models struggle to separate objects from transformations and reliably solve the object re-identification problem via in-context learning. What we need is a new approach that goes beyond preserving, re-learning, or adapting knowledge and instead leverages the compositional nature of continual learning problems by separating task-specific information from universal mechanisms.

\begin{figure*}[t]
    \includegraphics[width=\textwidth]{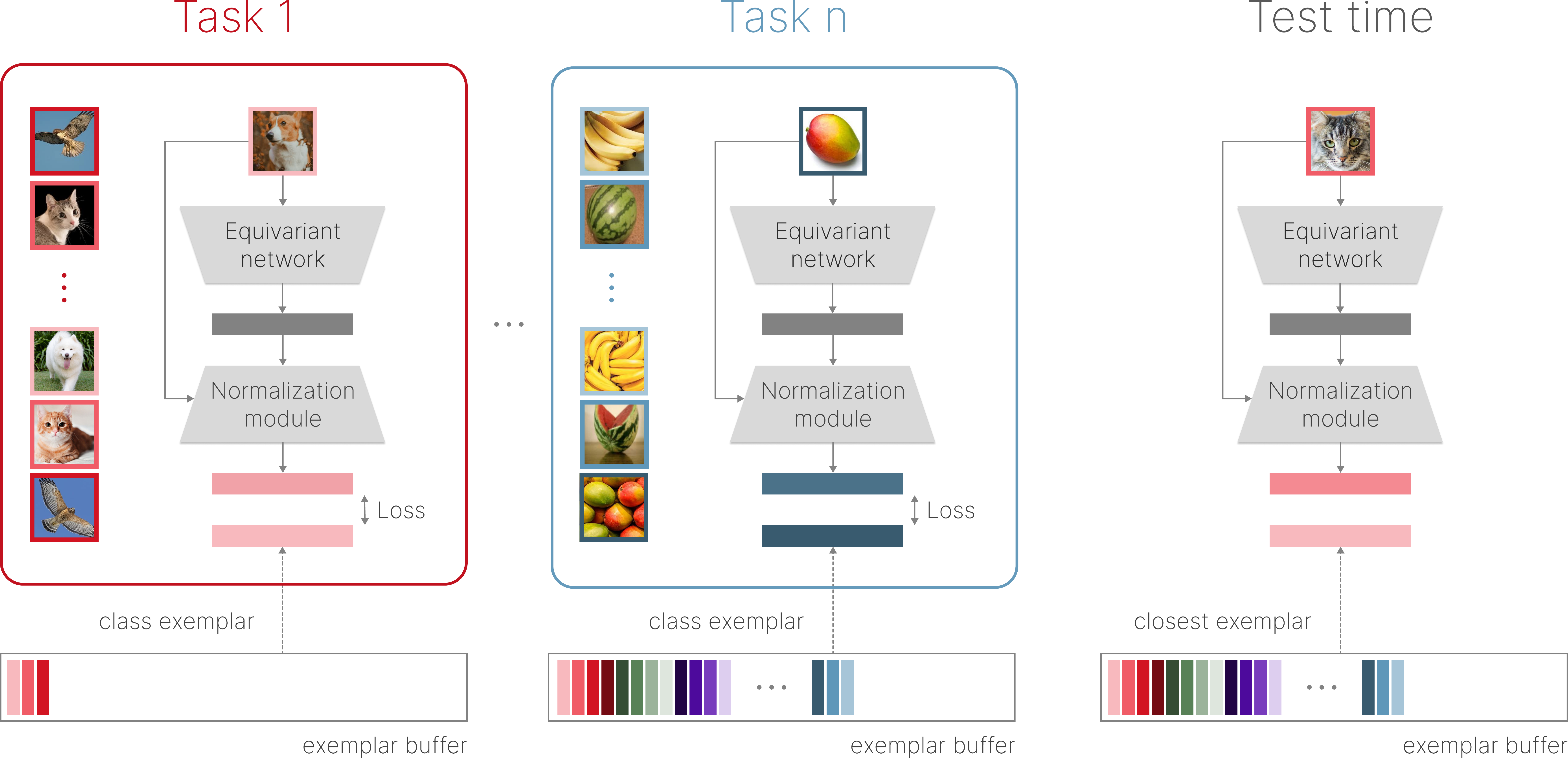}
    \caption{Schematic illustration of the conceptual \acrfull{dcl} framework with two modules transferred across tasks: 1)~the canonicalization network, consisting of an equivariant module that estimates the parameters of a transformation and a normalization module that applies this transformation to the input, and (2)~a buffer that stores class-specific exemplars. The canonicalization network is trained continually to map each image to its exemplar. At test time, we return the label of the exemplar closest to the normalized input.}
    \label{fig:fig1}
\end{figure*}

As noted by \cite{bower_catastrophic_1989}, catastrophic forgetting is caused by \emph{destructive model updates}, where adjusting model parameters through gradient descent to minimize the cost function of the current task impairs performance on past tasks. Inspired by this, we propose a novel paradigm for continual learning centered on separating memorization from generalization. We hypothesize that catastrophic forgetting can be minimized by decoupling two objectives: \emph{memorizing} class-specific information and \emph{learning} general mechanisms that transfer well across tasks. We aim to reduce destructive updates by having separate update procedures for the memory buffer and the generalization model. This allows us to expand, maintain, or prune class-specific knowledge in the memory buffer while continuously training the generalization model. By learning universal transformations, we avoid destructive gradient updates and efficiently accumulate knowledge over time. We call this approach \emph{\acrfull{dcl}} \footnote{Please note the difference from \emph{Disentangled Representation Learning}.}. Section \ref{subsec:model} describes a simple implementation of \acrshort{dcl}. It consists of an exemplar buffer that stores a single exemplar per class, an equivariant network that learns to regress parameters of an affine transform mapping any input to its exemplar, and a normalization module that applies the predicted affine transformation to the input.

\paragraph{Contributions} We summarize the most important contributions of this work below:
\begin{itemize}
    \item We introduce \acrfull{ids}, an open-source tool for generating continual classification and disentanglement benchmarks consisting of any number of unique tasks.
    \item We show that all major continual learning methods break down on a simple benchmark created with \acrshort{ids}.
    \item We propose \acrfull{dcl}, a novel approach to continual learning based on separating explicit memory edits from gradient-based model updates.
    \item We implement a proof of concept of \acrshort{dcl} and demonstrate it can efficiently learn over hundreds of tasks and perform open-set classification and zero-shot generalization.
\end{itemize}
\section{Motivation: Three issues with class-incremental continual learning}
\label{sec:primer}

\subsection{Benchmarking}
Continual learning datasets are typically limited to just a few tasks and at most a few hundred classes. In contrast, humans can learn and recognise countless novel objects throughout their lifetime. We argue that we should focus more on scaling the number of tasks in our benchmarks. We show that when tested over hundreds of tasks, standard methods inevitably fail: the impact of regularization decays over time, adding more parameters quickly becomes impractical, and replaying old samples eventually becomes ineffective under a constant computational budget. Moreover, to tackle individual sub-problems in continual learning, such as the influence of task similarity on forgetting, the role of disentangled representations, and the impact of hard task boundaries, we need to flexibly create datasets that allow us to isolate these issues. We should also move away from static training and testing stages and embrace streaming settings where the model can be evaluated at any point.

These observations motivated us to create a novel evaluation protocol. Taking inspiration from object-centric disentanglement libraries \citep{locatello2019challenging,Gondaletal19}, \acrshort{ids} allows for procedurally generating virtually infinite streams of data while retaining full control over the number of tasks, their difficulty and respective similarity, and the nature of boundaries between them.

\subsection{Invariant representations}
Continual learning methods are usually benchmarked on class-incremental setups, where a classification problem is split into several tasks to be solved sequentially \citep{gido2022three}. Note that the classification learning objective is \emph{invariant} to identity-preserving transformations of the object of interest, such as rigid transformations, deformations, change of lighting, or perspective projection. Unsurprisingly, the most successful discriminative learning architectures, from AlexNet \citep{krizhevsky2012imagenet} through ResNet \citep{he2016deep} to Vision Transformer \citep{dosovitskiy2020image}, learn only features relevant to the classification task and discard valuable information about universal transformations, symmetries, and compositions \citep{tishby2015deep, higgins2022symmetry}. By doing so, they entangle specific class information with knowledge about generalization mechanisms and represent both in the model's weights. When a new task arrives, there is no clear way to update these separately. We argue that transferring a purely discriminative model across tasks is not conducive to positive forward or backward transfer.

In this paper, we reframe the problem by recognizing that information about identity-preserving transformations, typically discarded, is crucial for transfer across tasks. For instance, changes in illumination affect objects of various classes similarly. Understanding this mechanism can lead to better generalization on future classes. Consequently, we propose that modeling these transformations is key to achieving positive forward and backward transfer in continual classification. Symmetry transformations, or equivariances, provide a structured framework for this modeling, which we elaborate on in the subsequent section.

\subsection{Pre-trained models}
Continual adaptation promises to combine the impressive generalization capabilities of large vision-language models trained on extensive amounts of data with the flexibility required for continual learning. While these methods achieve impressive results on many continual learning benchmarks, we argue that their strength comes primarily from their powerful backbone and its ability to solve individual tasks without much adaptation. \cite{zhou2023revisiting} recently demonstrated that freezing the model and classifying with the average embedding of each class outperforms sophisticated prompt tuning methods such as \acrlong{l2p} \citep{wang2022learning} and DualPrompt \citep{wang2022dualprompt}. Similarly, \cite{panos2023first} show that performing adaptation in the first learning session followed by \acrlong{lda} on top of a frozen network offers competitive performance on standard benchmarks.

Evaluating new types of methods requires new benchmarks. As general models pre-trained on Internet-scale data become common, it is important for continual learning benchmarks to remain challenging, requiring genuine adaptation and generalization to previously unseen data. Considering the emergent zero-shot abilities of foundation models, we should focus less on achieving competitive accuracy on existing benchmarks and more on re-thinking the desiderata, objectives, and metrics we use to design and evaluate continual learning systems. We agree with \cite{chollet2019measure} that the measure of intelligence is not only skill, but the speed of skill acquisition. We believe that new continual learning benchmarks should emphasize sample efficiency, positive forward and backward transfer, zero-shot and few-shot generalization, and open-ended learning. We see synthetic data as a perfect tool to measure progress in these areas.
\section{Methods}
\label{sec:method}
In this section, we describe in more detail the two main contributions of this work: a software package for generating continual learning benchmarks and a conceptual continual learning framework along with an example implementation. We would like to emphasize that our approach serves as a baseline and a proof of concept, showcasing the potential of \acrshort{dcl}, and is not intended as a practical method for general use. We see it as a first step towards efficient continual learning techniques that match humans in the ability to quickly generalize from very limited data and efficiently decompose \emph{any} classification problem into specific features that need to be explicitly memorized and general mechanisms that need to be learned.

\begin{figure}[t]
  \centering
  \includegraphics[width=\linewidth]{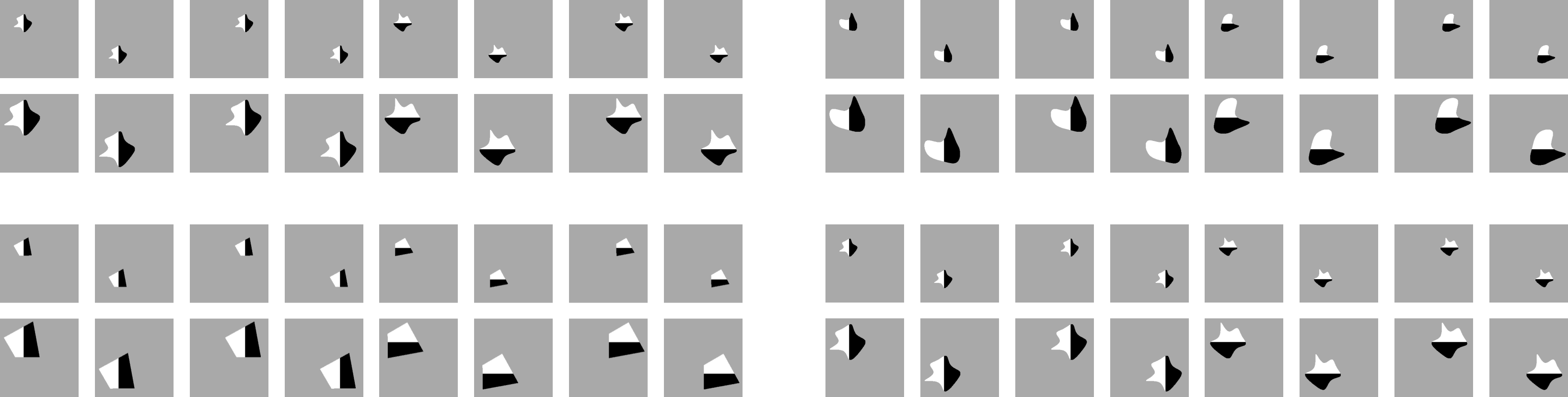}
  \caption{A batch of images from the Infinite dSprites dataset containing four distinct shapes. Each shape is shown in all combinations of four factors of variation (horizontal and vertical position, orientation, and scale) with two possible values per factor.}
  \label{fig:idsprites}
\end{figure}

\subsection{Infinite dSprites}
\label{subsec:idsprites}
\acrfull{ids} is a software framework designed for the easy creation of arbitrarily long continual learning benchmarks. A single \acrshort{ids} benchmark consists of $T$ tasks, where each task is an $n$-fold classification of procedurally generated shapes. Similar to dSprites, each shape is observed in all possible combinations of the following \glspl{fov}: color, scale, orientation, horizontal position, and vertical position. \Cref{fig:idsprites} shows an example batch of images with four \glspl{fov} and two values per factor (in general, our implementation allows for arbitrary granularity). The \emph{canonical form} corresponds to a scale of 1, an orientation of 0, and horizontal and vertical positions of 0.5. We only use a single color in our experiments for simplicity and to save computation.

The shapes are generated by first randomly sampling the number of vertices from a discrete uniform distribution over a closed integer interval $\llbracket a, b\rrbracket$, then constructing a regular polygon on a unit circle, randomly perturbing the polar coordinates of each vertex, and finally connecting the perturbed vertices with a closed spline of an order randomly chosen from $\{1, 3\}$. All shapes are then scaled and centered so that their bounding boxes and their centers of mass align in the canonical form. We also make orientation identifiable by painting one half of the shape black.

The number of tasks, the number of shapes per task, the vertex number interval, the exact \gls{fov} ranges, and the parameters of noise distributions for radial and angular coordinates are set by the user, providing the flexibility to control the length, structure, and difficulty of the benchmark. The framework also provides access to the ground truth values of the individual \glspl{fov}. \acrshort{ids} is a pip-installable Python package that we hope will unlock new research directions in continual classification, transfer learning, and continual disentanglement.

\subsection{Disentangled continual learning}
\label{subsec:model}
As mentioned earlier, in order to efficiently solve \acrshort{ids}, we need to clearly distinguish between the generalization mechanism that needs to be learned and the class-specific information that has to be memorized. We start by observing that human learning is likely characterized by such separation. Take face recognition, for example. A child can memorize the face of its parent but may still get confused by an unexpected transformation, as evidenced by countless online videos of babies failing to recognize their fathers after a shave. Once we learn the typical identity-preserving transformations that a face can undergo, we only need to memorize the particular features of any new face to instantly generalize over many transformations, such as facial expressions, lighting, three-dimensional rotation, scale, and perspective projection. Note that while we encounter new faces every day, these transforms remain consistent and affect every face similarly. Indeed, this fundamental property of the physical world makes generalization possible. As \cite{taylor2021we} aptly note, “to successfully generalize information for appropriate use in novel situations, learned information should reflect stable features of the environment and discard idiosyncratic details of individual experiences.”

Inspired by this observation, we aim to disentangle generalization from memorization by explicitly separating the learning module from the memory buffer in our model design. The memory buffer stores a single exemplar image of each encountered shape. We assume these are provided by an oracle throughout training, but it would be possible to bootstrap the buffer with a few initial exemplars. The equivariance learning module is a neural network designed to encapsulate the general transformations present in the data by learning to \textit{canonicalize} each input, i.e. transform it to the canonical form. At test time, each input image is canonicalized and then compared to the stored exemplars. This approach draws inspiration from prototype and exemplar-based models of categorization in neuroscience, which operate on the premise that the brain compares the current stimulus with representations of all pertinent categories and selects the one perceived as most similar \citep{bowman2018abstract, nosofsky2000exemplar}.

\subsubsection{Implementation and training objective}
At each task, we observe a training set $D_n$ of triplets, each comprising an image $x$, its generative factors $y$, and a class label $c$. Since we are tackling the class-incremental scenario \citep{gido2022three}, we do not have access to task labels. However, we do assume that for each class we are given an exemplar: a single image showing the shape in the canonical form. Whenever a new class is encountered, we add its exemplar to the memory buffer. We continually train a neural network to regress the parameters of a two-dimensional affine transformation that maps each image to its class exemplar. We supervise the network directly with MSE loss on the transformation parameters, since we can access the generative factors for each input image and calculate the ground truth transform. At test time, we use this network to canonicalize images of previously unseen shapes and compare them with exemplars stored in the buffer. Each image is then classified as belonging to the class of its nearest exemplar. Please see \cref{alg:dcl_train} and \cref{alg:dcl_test} for details of the implementation.

\subsubsection{Discussion}
The disentangled learning approach has a number of advantages. First, by learning transformations instead of class boundaries, we reformulate a challenging class-incremental classification scenario as a domain-incremental \gls{fov} regression learning problem \citep{gido2022three}. Since the transformations affect every class in the same way, they are easier to learn in a continual setting. We show that this approach is not only less prone to forgetting but exhibits significant forward and backward transfer. In other words, the knowledge about regressing \glspl{fov} is efficiently accumulated over time. Second, the exemplar buffer is a fully explainable representation of memory that can be explicitly edited: we can easily add a new class or completely erase a class from memory by removing its exemplar. Finally, we show experimentally that our method generalises instantly to new shapes with just a single exemplar and works reliably in an open-set classification scenario.

\begin{figure}[htbp]
\begin{minipage}[t]{0.48\linewidth}
\begin{algorithm}[H]
\caption{Online \acrshort{dcl} on \acrshort{ids} (train)}
\label{alg:dcl_train}
\begin{algorithmic}[1]
\State \textbf{Input}: Data $D=\{D_{n}\}_{n=1}^M, D_n=\{(x_{i}, y_{i}, c_{i})\}_{i=1}^{N_{n}}$
\State \textbf{Input}: Exemplars dictionary $E$
\State \textbf{Initialize}: Memory buffer dictionary $M$
\State \textbf{Initialize}: Canonicalization network $f_{\theta}$
\For{each task $n$}
    \For{each $(x_i, y_i, c_i)$ in $D_n$}
        \If {$c_i$ \textbf{not in} $M$}
            \State $M[c_i] \gets E[c_i]$
        \EndIf
        \State $T \gets \Call{ComputeTransformation}{y_i}$
        \State $\mathcal{L} \gets \Call{MSE}{f_{\theta}(x_i), T}$
        \State $\theta \gets \Call{OptimizerStep}{\mathcal{\theta, L}}$
    \EndFor
\EndFor
\end{algorithmic}
\end{algorithm}
\end{minipage}%
\hfill
\begin{minipage}[t]{0.48\linewidth}
\begin{algorithm}[H]
\caption{Online \acrshort{dcl} on \acrshort{ids} (test)}
\label{alg:dcl_test}
\begin{algorithmic}[1]
\State \textbf{Input}: Data point to classify $x$
\State \textbf{Input}: Memory buffer dictionary $M$
\State \textbf{Input}: Trained canonicalization network $f_{\theta}$
\State \textbf{Initialize}: Minimum distance $d_{\text{min}} \gets \infty$
\State \textbf{Initialize}: Class label $c \gets \text{None}$

\State Canonicalize $\bar{x} \gets f_{\theta}(x)$
\For{each $c_i, \hat{x}_i$ in $M$}
\State $d \gets \Call{MSE}{\hat{x}_i, \bar{x}}$
    \If{$d < d_{\text{min}}$}
        \State $d_{\text{min}} \gets d$
        \State $c \gets c_i$
    \EndIf
\EndFor
\State \textbf{Output}: Class label $c$
\end{algorithmic}
\end{algorithm}
\end{minipage}
\end{figure}

\section{Related work}
\label{sec:related_work}

\subsection{Continual learning}
Continual learning literature typically focuses on catastrophic forgetting in supervised classification. \textit{Parameter isolation} methods use dedicated parameters for each task by periodically extending the architecture while freezing already trained parameters \citep{rusu_progressive_2016} or by relying on isolated sub-networks \citep{fernando_pathnet_2017}. \textit{Regularization} approaches aim to preserve existing knowledge by limiting the plasticity of the network. Functional regularization methods constrain the network output through knowledge distillation \citep{li2017learning} or by using a small set of anchor points to build a functional prior \citep{pan_continual_2020, titsias_functional_2020}. Weight regularization methods \cite{zenke_continual_2017} directly constrain network parameters according to their estimated importance for previous tasks. In particular, \gls{vcl} by \cite{nguyen_variational_2017} derives the importance estimate by framing continual learning as sequential approximate Bayesian inference. Most methods incorporate regularization into the objective function, but it is also possible to implement it using constrained optimization \cite{lopez-paz_gradient_2017, aljundi_gradient_2019, hess2023two, kao2021natural}. Finally, \textit{replay} methods \citep{rebuffi_icarl_2017, chaudhry_tiny_2019, isele_selective_2018, rolnick_experience_2019} retain knowledge through rehearsal. When learning a new task, the network is trained with a mix of new samples from the training stream and previously seen samples drawn from the memory buffer. A specific case of this strategy is \textit{generative replay} \citep{shin_continual_2017, atkinson_pseudo-recursal_2018}, where the rehearsal samples are produced by a generative model trained to approximate the data distribution for each class. Finally, continual adaptation approaches rely on a large pre-trained model that is then 

\subsection{Benchmarking continual learning}
Established continual learning benchmarks primarily involve splitting existing computer vision datasets into discrete, non-overlapping segments to study continual supervised classification.
Notable examples in this domain include split MNIST \citep{zenke_continual_2017}, split CIFAR \citep{zenke_continual_2017}, and split MiniImageNet \citep{chaudhry_tiny_2019,Aljundi19Online}, along with their augmented counterparts, such as rotated MNIST \citep{lopez-paz_gradient_2017}, and permuted MNIST \citep{kirkpatrick_overcoming_2017}. Contributions from \citet{lomonaco_core50_2017}, \citet{verwimp2023clad} and \citet{Roady_2020_Stream51} have enriched the field with dataset designed specifically for continual learning, such as CORe50, CLAD, and Stream-51, which comprise temporally correlated images with diverse backgrounds and environments. More recently, \cite{lesort2023challenging} showed how scaling up continual learning benchmarks can reveal new insights about knowledge accumulation and catastrophic forgetting. Similar to \acrshort{ids}, their experimentation framework offers the capability to create any number of tasks, although in contrast to our procedural approach, they construct the tasks by randomly recombining a finite set of classes from an existing dataset to investigate the effect of data reocurrence.
\section{Experiments}
\label{sec:exp}
In this section, we evaluate standard continual learning methods and our disentangled learning framework on a benchmark generated using \gls{ids}. The benchmark consists of 500 classification tasks. For each task, we randomly generate 10 shapes and create an image dataset showing each shape in all combinations of 4 \glspl{fov} with 5 possible values per factor, resulting in 6,250 samples per task, which we then randomly split into training, validation, and test sets with a 6000:150:100 ratio. After training on each task, we report the average test accuracy on all tasks seen so far. To ensure a reliable comparison, we use a ResNet-18 backbone for every method, except for Learning to Prompt, which uses a Vision Transformer  \citep{he2016deep, dosovitskiy2020image}. We make sure that all models are trained until convergence. We aim to understand the performance of standard continual learning methods and \acrshort{dcl} over a long time horizon and make sure that \acrshort{ids} strikes the right balance between simplicity and sufficient complexity to pose a challenge for existing methods.

\subsection{Regularization methods}
\label{subsec:exp:reg}
We compare our approach to standard regularization methods: \gls{lwf}, \gls{si}, and \gls{ewc} \citep{li_learning_2017, zenke_continual_2017, kirkpatrick_overcoming_2017}. We use implementations from Avalanche, a continual learning library by \cite{lomonaco2021avalanche}. We provide details of the hyperparameter choice in the supplementary material. As shown in \cref{fig:regularization}, such regularization methods are ill-equipped to deal with long horizon class-incremental learning scenario and their performance deteriorates rapidly after only 10 tasks. This finding has been previously described by \cite{lesort2019regularization} and \cite{lomonaco2020rehearsal}, and theoretically supported by \cite{knoblauch2020optimal}.

\subsection{Replay-based methods}
\label{subsec:exp:rep}
\begin{figure}[t]
    \centering
    \begin{minipage}{0.48\textwidth}
        \centering
        \includegraphics[width=\textwidth]{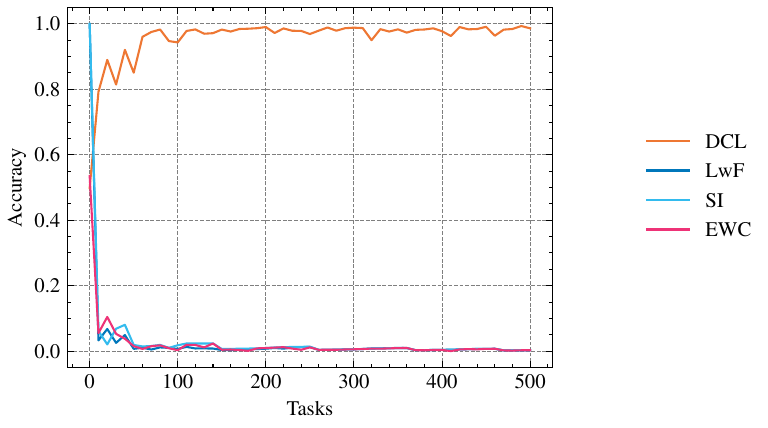} 
        \caption{Average test accuracy on all past tasks for \acrlong{dcl} and standard regularization methods: \acrlong{lwf}, \acrlong{si}, and \acrlong{ewc}.}
        \label{fig:regularization}
    \end{minipage}\hfill
    \begin{minipage}{0.48\textwidth}
        \centering
        \includegraphics[width=\textwidth]{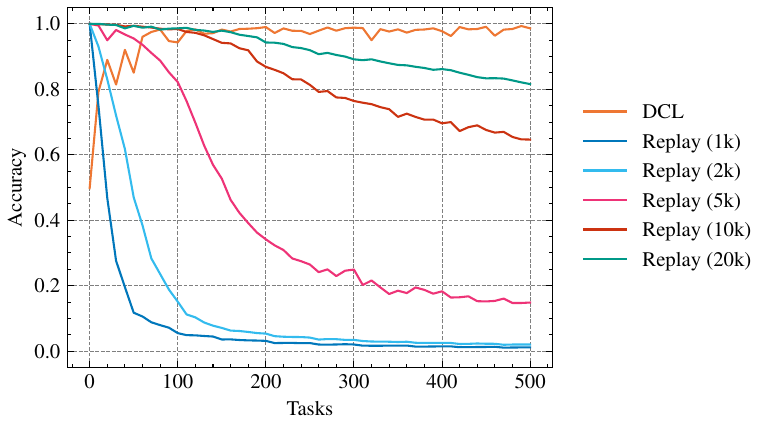} 
   \caption{Average test accuracy on all past tasks for \acrlong{dcl} and Experience Replay with different buffer sizes (in number of images). Note that after 500 tasks \acrshort{dcl} only stores 5,000 images.}
        \label{fig:experience_replay}
    \end{minipage}
\end{figure}

Rehearsal can be a viable strategy to maintain high accuracy over hundreds of tasks, but unless the buffer size is bounded, its memory footprint grows rapidly over time. In this section, we investigate the effect of maximum buffer size on performance for standard experience replay with reservoir sampling. While there are replay-based methods that improve on this baseline, we are interested in investigating the fundamental limits of rehearsal over long time horizons and strip away the confounding effects of data augmentation, generative replay, sampling strategies, etc. As seen in \cref{fig:experience_replay}, even with a buffer of 20,000 images, the accuracy of experience replay decreases after only a few dozen tasks. After 500 tasks, the buffer contains merely a few samples per class and the accuracy drops considerably, despite using double the compute resources of alternative methods. This is consistent with previous findings of \cite{wang2022dualprompt}, who observe the performance of replay-based methods deteriorates as the buffer size shrinks. Conversely, the accuracy degrades for a set buffer size as the model encounters more tasks. It would be possible to combat forgetting by further expanding the buffer size, but given a limited computational budget, the model would still encounter only a fraction of the buffer during training. This fundamental limitation make replay-based methods impractical when learning over a large number of tasks. For completeness, we provide additional comparisons to \acrlong{agem} \citep{chaudhry2018efficient} and \acrlong{der} \citep{buzzega2020dark} in \cref{subsec:app:rep}.

\subsection{Prompt-based methods}
\label{subsec:exp:pre}
With the advent of large vision-language models, parameter-efficient fine-tuning emerged as a popular approach to continual learning. Typically, it involves pre-training a large model on a large dataset like ImageNet-21k \citep{russakovsky2015imagenet, ridnik2021imagenet} and continually adapting a smaller set of parameters in the hope of combining the power of a general-purpose model with the plasticity of prompt tuning \citep{hu2021lora, jia2022visual}. These methods achieve impressive results on natural image benchmarks, such as Split-CIFAR or Split-ImageNet, whose categories largely overlap with the training data of the original foundation model. Nonetheless, in many real-world problems of interest such as remote sensing, astronomy, medical imaging, or specialised vision systems, the data follows completely different distributions. Adapting to domains that are not covered by the pre-training datasets is more difficult and prone to catastrophic forgetting. To demonstrate it, we test \acrfull{l2p}, a prototypical continual prompt tuning method introduced by \cite{panos2023first} on our synthetic dataset. As shown in \cref{fig:res:pretrained}, the performance of the 86M parameter ViT-B/16 model on this simple dataset rapidly deteriorates. With this experiment, we demonstrate that despite it synthetic nature, \acrshort{ids} still presents a considerable challenge for continual learning methods based on pre-trained models.

\subsection{Foundation models}
\label{subsec:exp:gpt}
\begin{figure}[t]
    \centering
        \begin{minipage}{0.48\textwidth}
        \centering
        \includegraphics[width=\textwidth]{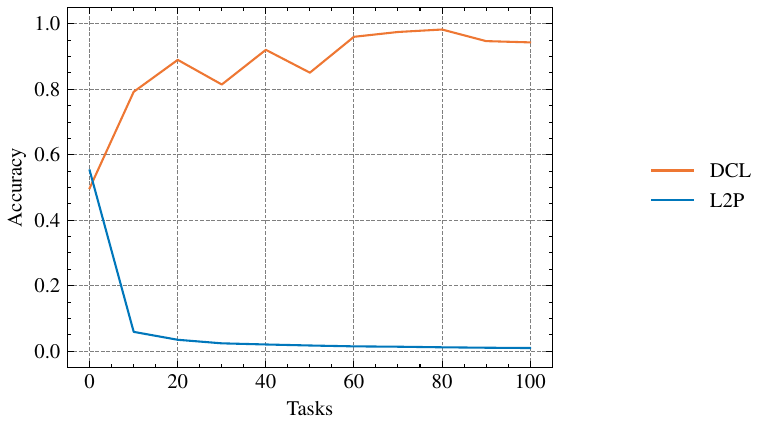}
        \caption{Cumulative test accuracy of \acrlong{dcl} and \acrfull{l2p}. Note that even after training for 20 epochs on the first task, the accuracy of \acrshort{l2p} is below 60\%, suggesting that prompt tuning methods have limited adaptability to domains that differ a lot from the pre-training distribution. We run a 100 tasks because of computational requirements of \acrshort{l2p}.}
        \label{fig:res:pretrained}
    \end{minipage}\hfill
    \begin{minipage}{0.48\textwidth}
        \centering
        \includegraphics[width=\textwidth]{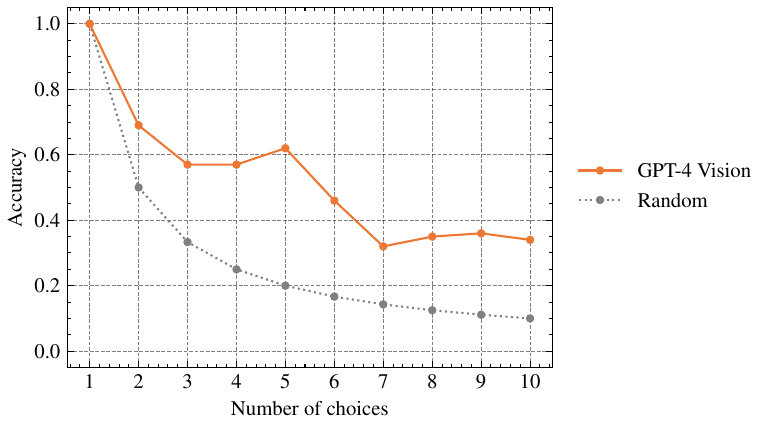}
        \caption{Average accuracy of the GPT-4 Vision model on the multiple choice shape re-identification task. The model is presented with a query shape in a random position, orientation, and scale, and asked to pick the correct exemplar out of n possible choices. The accuracy is calculated over 100 trials. Accuracy of a random guess is plotted for reference.}
        \label{fig:res:gpt}
    \end{minipage}\hfill
\end{figure}

Multimodal foundation models can perform an impressive range of computer vision tasks zero-shot or via in-context learning. Their capabilities include classification, visual question answering, object detection, and optical character recognition. In this section, we test whether GPT-4, a state of the art foundation model introduced by \cite{achiam2023gpt} is already able to solve the shape re-identification task that lies at the core of \acrshort{ids}. To this end, we present the model with a query shape in a random position, orientation, and scale. We then show multiple shapes in the canonical form and ask it to pick the exemplar matching the query. \Cref{fig:res:gpt} shows the average accuracy over 100 multiple-choice questions as a function of the number of possible answers. While the model performs better than chance when there are only a few answers to choose from, we conclude that GPT-4 zero-shot object re-identification capabilities are not yet at a level where it could reliably solve our simple benchmark. The details of the experiment including the exact prompt are provided in \cref{subsec:app:gpt}.

\subsection{Disentangled continual learning}
\label{subsec:exp:dcl}
The particular implementation of Disentangled Continual Learning for Infinite dSprites is so effective because it separates memorization from generalization by introducing a strong inductive bias, constraining the class of symmetries expected in the data to two-dimensional affine transformations. The crucial advantage of this approach is that continually regressing the parameters of these transformations is not prone to catastrophic forgetting. We speculate this is because the underlying objective remains constant even as new shapes appear. The network needs to learn a general representation and find an effective algorithm for estimating the factors of variation: scale, orientation, and position. Since each shape is observed in numerous combinations of these factors, relying on a shape-specific shortcut would fail, leading the network to learn a more general solution. Furthermore, this solution is refined as new tasks appear, indicating that naive fine-tuning on the FoV regression task in a domain-incremental setting is enough to achieve knowledge accumulation.

We believe this finding can be explained by two main factors. First, gradient-based optimization can overcome catastrophic forgetting and demonstrate knowledge accumulation when data reoccurs over a long sequence, as shown recently by \cite{lesort2023challenging}. While each shape in idSprites is unique, there is perhaps enough similarity within types of shapes to lead to improved FoV regression, thereby boosting classification accuracy. Second, because the underlying task of FoV regression remains constant, it is reasonable to expect the network to exhibit less feature forgetting. According to a study by \cite{hess2023knowledge}, this consistency is likely to contribute to improved knowledge accumulation.

\subsection{Do we need equivariance?}
\label{subsec:exp:eqv-inv}
To demonstrate further that learning an equivariant representation is the key to achieving effective continual learning within our framework, we compare equivariant and invariant learning directly. Our baseline for invariant representation learning is based on SimCLR \citep{chen2020simple}, a simple and effective contrastive learning algorithm that aims to learn representations invariant to data augmentations. To adapt SimCLR to our problem, we introduce two optimization objectives. The first objective pulls the representation of each training point towards the representation of its exemplar while repelling all other training points. The second objective encourages well-separated exemplar representations by pushing the representations of all exemplars in the current task away from each other. We observed that the first training objective alone is sufficient, but including the second loss term speeds up training. For each task, we train the baseline until convergence. At test time, the class labels are assigned through nearest neighbor lookup in the representation space. Similar to our method, we store a single exemplar per class.

\begin{figure}[t]
    \centering
    \begin{minipage}{0.48\textwidth}
        \centering
        \includegraphics[width=\textwidth]{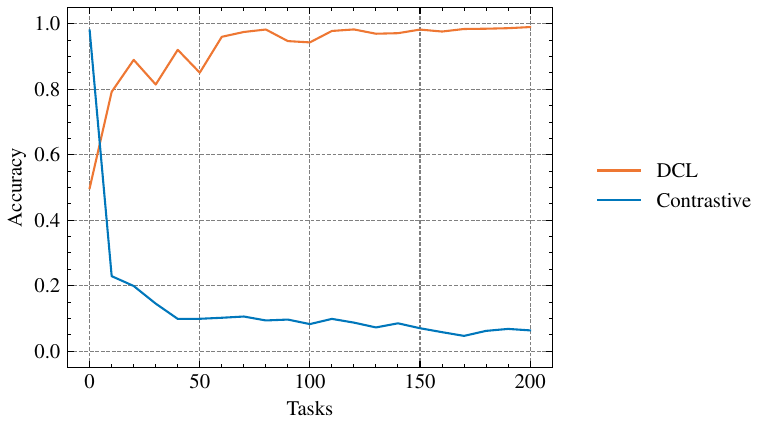}
        \caption{Cumulative test accuracy of \acrlong{dcl} and an invariant contrastive learning baseline. The contrastive baseline stays well above the chance level after observing a thousand shapes.}
        \label{fig:res:contr-vs-equiv}
    \end{minipage}\hfill
    \begin{minipage}{0.48\textwidth}
        \centering
        \includegraphics[width=\textwidth]{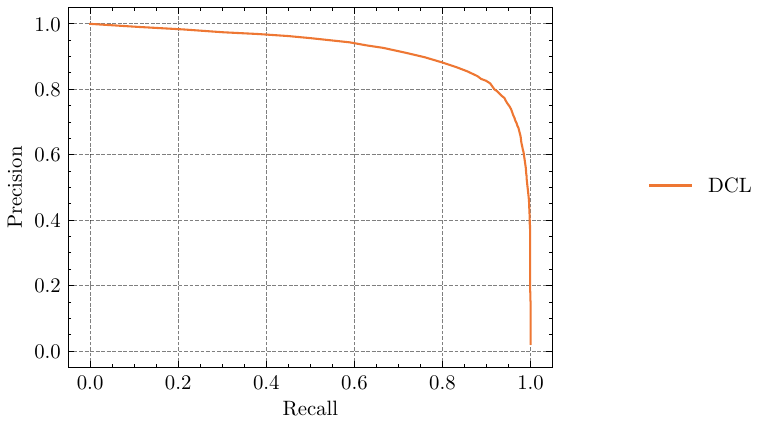}
        \caption{Precision-recall curve for open-set classification with \acrlong{dcl}. Each point on the curve corresponds to a different value of the classification threshold $\sigma$.}
        \label{fig:res:corresp}
    \end{minipage}\hfill
\end{figure}

\Cref{fig:res:contr-vs-equiv} shows test accuracy for both methods over time. The performance of the contrastive learning baseline decays over time, but not as rapidly as naive fine-tuning. Note that in contrast to our method, invariant learning could benefit from storing more than one exemplars per class. The supplementary material provides an exact formulation of the contrastive objective and implementation details.

\subsection{One-shot generalization}
\label{subsec:exp:one-shot}
\begin{figure}[htbp]
    \centering
    \begin{minipage}{0.48\textwidth}
        \centering
        \includegraphics[width=\textwidth]{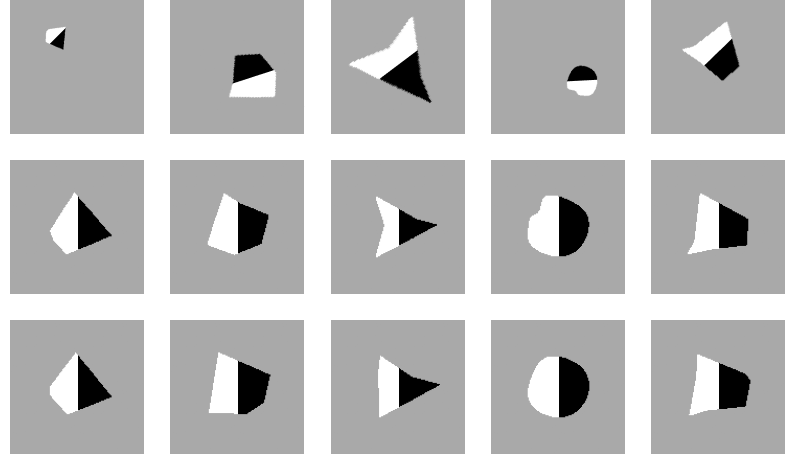}
        \caption{Some of the incorrectly classified unseen shapes in the open-set recognition task. Top row: inputs to the equivariant network. Middle row: outputs of the normalization module. Bottom row: closest training exemplars.}
        \label{fig:res:corresp2}
    \end{minipage}\hfill
    \begin{minipage}{0.48\textwidth}
        \centering
        \captionof{table}{One-shot generalization accuracy of \acrlong{dcl} on unseen shapes after training for 200 tasks (10 shapes per task). In the standard variant, the buffer is cleared before adding novel exemplars. In the more challenging generalized variant, the exemplars for unseen shapes are appended to the buffer. Keeping the training exemplars in the buffer, as well as including more test classes result in slightly worse classification accuracy. }
        \label{tab:one-shot}
        \newcolumntype{Y}{>{\raggedleft\arraybackslash}X}
        \begin{tabularx}{\textwidth}{@{}lYYYY@{}}
        \toprule
        \multirow{2}{*}{} & & \multicolumn{3}{r}{Novel classes added}\vspace{2mm} \\
        & & 10 & 100 & 1000 \\
        \midrule
        Standard     & & 1.00  & 0.98 & 0.95 \\
        Generalized  & & 0.96 & 0.95 & 0.93 \\
            \bottomrule
        \end{tabularx}
    \end{minipage}
\end{figure}

To evaluate whether the learned regression network can generalize to unseen classes, we perform a one-shot learning experiment. Here, the model is asked to classify transformed versions of shapes it had not previously encountered. Since the returned class label depends on the exemplars in the buffer, we consider two variants of the experiment, corresponding to generalized and standard one-shot learning. In the first one, we keep the training exemplars in the buffer and add new ones. In the second, the buffer is cleared before including novel exemplars. We also introduce different numbers of test classes. The classification accuracies are presented in Table~\ref{tab:one-shot}. As expected, keeping the training exemplars in the buffer and adding more test classes makes the task harder. Nevertheless, the accuracy stays remarkably high, showing that the equivariant network has learned a correct and universal mechanism that works even for unseen shapes. This is the essence of our framework.

\subsection{Open-set classification}
\label{subsec:exp:have-i-seen}
Next, we investigate how well our proposed framework can detect novel shapes. This differs from the one-shot generalization task because we do not add the exemplars corresponding to the novel shapes to the buffer.
Instead of modifying the learning setup, we use a simple heuristic based on an empirical observation that our model can almost perfectly normalize any input---we classify the input image as unseen if we cannot find an exemplar that matches the normalized input significantly better than others.

Denoting the equivariant network output by $\hat{\theta}$ and the two best candidate exemplars by $c_1$ and $c_2$, we classify the test input as novel if $\|c_1-T_{\hat{\theta}}(x) \|_2^2 > \sigma \|c_2-T_{\hat{\theta}}(x) \|_2^2$. \Cref{fig:res:corresp} shows the precision-recall curve for different values of $\sigma$, with an overall area under curve of 0.92.
We also present qualitative results in \cref{fig:res:corresp2}.
\section{Discussion}
\label{sec:discussion}

In the last decade, continual learning research has made progress through parameter and functional regularization, rehearsal, and architectural strategies that mitigate forgetting by preserving important parameters or compartmentalizing knowledge.
As pointed out in a recent survey \citep{gido2022three}, the best performing continual learners are based on storing or synthesizing samples.
Such methods are typically evaluated on sequential versions of standard computer vision datasets such as MNIST or CIFAR-100, which often involve only a small number of learning tasks, discrete task boundaries, and fixed data distributions.
As such, the benchmarks do not match the lifelong nature of real-world learning tasks.

Our work is motivated by the hypothesis that state-of-the-art continual learners would inevitably fail when trained in a true lifelong fashion akin to humans. To test our claim, we use \acrlong{ids} to create a benchmark of procedurally generated shapes under affine transformations. To our knowledge, this is the first class-incremental continual learning benchmark that allows generating thousands of unique tasks. While we acknowledge the simplistic nature of our dataset, we believe any \emph{lifelong} learner must be able to solve \acrshort{ids} before tackling more complicated, real-world datasets. Nevertheless, our empirical findings highlight that standard methods are doomed to collapse and memory buffers only defer the ultimate end.

Updating synaptic connections in the human brain upon novel experiences does not interfere with the general knowledge accumulated throughout life. Inspired by this insight, we introduce \acrlong{dcl}, which decomposes the continual learning problem into (1) memorizing class-specific information relevant to the task and (2) sequentially training a network that models the general aspects of the problem that apply to all instances.
This separation enables explicitly updating class-specific information without destroying information pertinent to other classes and continually learning equivariant representations without catastrophic forgetting.
As demonstrated experimentally, a method implementing this spearation exhibits successful forward and backward transfer, one-shot generalization, and open-set recognition.

\paragraph{Limitations} With this work, we aim to bring a fresh perspective and chart a novel research direction in continual learning. To demonstrate our framework, we stick to a simple dataset and embed the correct inductive biases in our learning architecture. We acknowledge that when applied to natural images, our approach would suffer from a number of issues, which we list below, along with some mitigation strategies.

\begin{itemize}
    \item Real-world data does not come with perfect supervision signals, hindering the learning of equivariant networks. As a remedy, one might employ equivariant architectures as an inductive bias \citep{finzi2021practical} or weakly supervise the learning, e.g. with image-text pairs \citep{radford2021learning}.
    \item Obtaining canonical class exemplars for real-world data is not straightforward, which makes training the normalization network difficult. However, with a powerful enough re-identification capability, an arbitrary example of a class can serve as an exemplar.
    \item Even though the data generating process of many problems exhibits the compositional nature we model with \acrshort{ids}, it is not clear whether we can separate memorization and generalization for any continual learning problem. We plan to further investigate this question with more complex datasets.
\end{itemize}

\paragraph{Societal impact}
Similar to rehearsal method, \acrlong{dcl} stores data from past tasks. In case of personally identifiable information, this design decision might have privacy implications. However, in contrast to standard continual learning methods, the class-specific knowledge is stored only in the exemplar buffer and not in the weights of a neural network. As a result, removing the data pertaining to a given individual (in accordance for example with the right to erasure under GDPR) is much easier in our framework, as deleting it from the buffer is enough to guarantee no information is retained.

\clearpage
\paragraph{Acknowledgements}
This work was supported by the German Federal Ministry of Education and Research (BMBF): Tübingen AI Center, FKZ: 01IS18039A. This research utilized compute resources at the Tübingen Machine Learning Cloud, DFG FKZ INST 37/1057-1 FUGG. We thank the International Max Planck Research School for Intelligent Systems (IMPRS-IS) for supporting SD. This work was supported by the National Centre of Science (Poland) Grants No. 2020/39/B/ST6/01511 and 2022/45/B/ST6/02817.

\bibliography{main}
\bibliographystyle{collas2024_conference}

\clearpage
\appendix
\setcounter{section}{0}
\setcounter{figure}{0}
\FloatBarrier
\section{Appendix: Experimental details}
All models except for \acrshort{l2p} employ the same ResNet-18~\citep{he2016deep} backbone and are trained using the Adam optimizer~\cite{kingma2014adam} with default PyTorch~\citep{paszke2019pytorch} parameter values ($\lambda=0.001$, $\beta_{1}=0.9$, $\beta_{2}=0.999$). \acrshort{l2p} uses a ViT-B/16 \citep{dosovitskiy2020image} pretrained on ImageNet-21k \cite{ridnik2021imagenet} and fine-tuned on ImageNet 2012 \cite{russakovsky2015imagenet}.

\subsection{Regularization methods}
We ran a grid search to set the loss balance weight $\lambda_{o}$ in \gls{lwf} \cite{li_learning_2017} and the strength parameter $c$ in \gls{si} \cite{zenke_continual_2017}, but found the choice of hyperparameter did not influence the result. For the run shown in Figure \cref{fig:regularization}, we used $\lambda_{o}=0.1$ and $c=0.1$.

\subsection{Foundation models}
\label{subsec:app:gpt}
We use the GPT-4 Vision API to perform the multiple-choice experiment. The instruction prompt and the query shape are provided in separate messages, followed by a third message containing the images to pick from. We use the following prompt:

\begin{mdframed}
You will be shown a query image containing a black and white shape on a gray background. You will then be shown \{num\_choices\} images, one of which is the same shape as the query image, but rotated, translated, and scaled. Please select the image that matches the query image. Please only select one image. Please only
output a single number between 0 and \{num\_choices - 1\} (inclusive) indicating your choice.
\end{mdframed}

For each value of num\_choices between 2 and 10, we generate a 100 query shapes and corresponding answer sets and compute average accuracy on the multiple choice task.

\subsection{Contrastive baseline}
The optimization objective of the contrastive baseline consists of two components. The first one ensures that each sample in the batch is pulled towards its exemplar and pushed away from all the other samples in the mini-batch that belong to a different class. Using the terminology from \citep{chen2020simple}, a sample $x_{i}$ and its corresponding exemplar $\hat{x}_{i}$ constitute a \emph{positive pair}. Denoting the class of a sample as $c(x)$, the mini-batch size as $N$, and the representation of $x_{i}$ and $\hat{x}_{i}$ as $z_{i}$ and $\hat{z}_{i}$, respectively, the first component of the loss is:
\begin{equation}
l_{1}(x_{i}) = -\log \frac{\exp(sim(z_i, \hat{z_{i}}) / \tau)}{\sum_{k=1}^{N} \mathbbm{1}_{[c(k) \neq c(i)]} \exp(sim(z_i, z_k) / \tau)},
\end{equation}
where $sim(z_{i}, z_{j})$ is the cosine similarity between $z_{i}$ and $z_{j}$.

The second component encourages well-separated exemplar representations by pushing apart the representations of all the exemplars in the current mini-batch. Denoting the number of distinct classes in the mini-batch as $C$, we have:
\begin{equation}
l_{2}(x_{i}) = -\log \frac{\exp(1 / \tau)}{\sum_{k=1}^{C} \mathbbm{1}_{[k \neq i]} \exp(sim(\hat{z_i}, \hat{z_k}) / \tau)}.
\end{equation}
The final mini-batch loss is $\mathcal{L}=\sum_{i=1}^{N}l_{1}(x_{i}) + l_{2}(x_{i})$.

\clearpage
\section{Appendix: Additional experiments}
\subsection{Replay-based methods}
\label{subsec:app:rep}
\begin{figure}[htpb]
    \centering
    \begin{minipage}{0.48\textwidth}
        \centering
        \includegraphics[width=\textwidth]{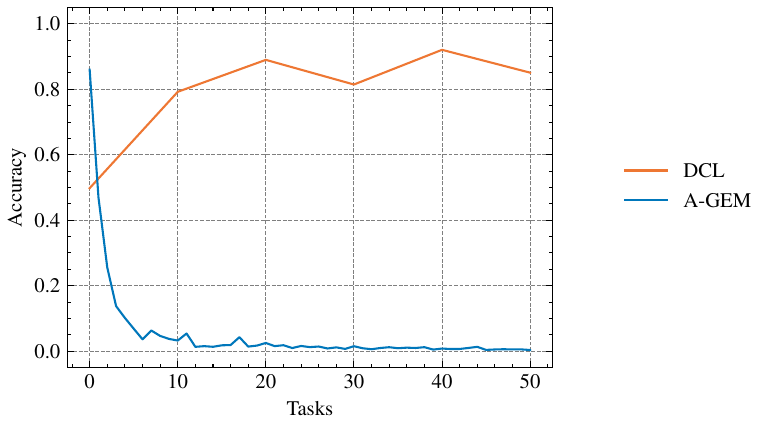}
        \caption{Average test accuracy on all past tasks for \acrlong{dcl} and \acrfull{agem}.}
        \label{fig:res:agem}
    \end{minipage}\hfill
        \begin{minipage}{0.48\textwidth}
        \centering
        \includegraphics[width=\textwidth]{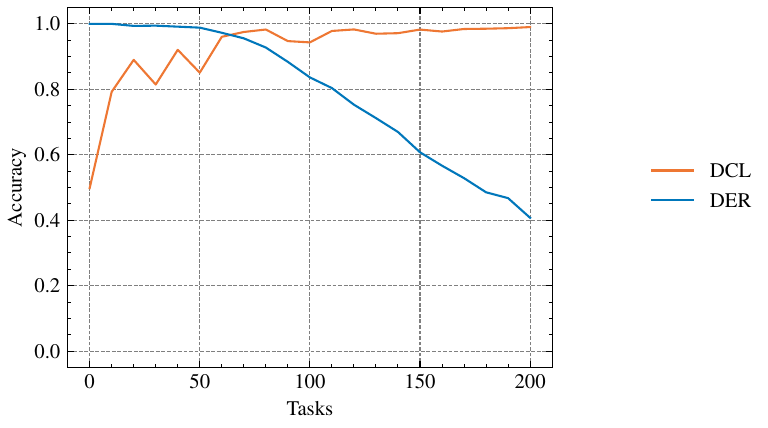}
        \caption{Average test accuracy on all past tasks for \acrlong{dcl} and \acrfull{der} with a buffer of 5k images.}
        \label{fig:res:der}
    \end{minipage}\hfill
\end{figure}

In addition to Experience Replay results presented in the main paper, we compare \acrshort{dcl} to two other baselines that use a rehearsal buffer. \acrfull{agem}, proposed by \cite{chaudhry2018efficient}, maintains an episodic memory for each task seen so far. When optimising for the current task, the model is prevented from decreasing the loss on the episodic memories through inequality constraints. Even though A-GEM is a more efficient version of the original \acrfull{gem}, we found it to be very computationally expensive and therefore only run the experiment for 50 tasks. The results are shown in \cref{fig:res:agem}.

\acrfull{der}, introduced by \cite{buzzega2020dark} is a replay-based method that promotes consistency with the past by matching the logits of the network sampled throughout the optimization trajectory. As shown in \cref{fig:res:der}, \acrshort{der} with a buffer of 5,000 images outperforms Experience Replay with the same buffer size (see \cref{fig:experience_replay}), but still deteriorates over time, reaching the accuracy of 40\% after only 200 tasks.

\subsection{Online vs. offline}
\label{subsec:exp:onl-off}
\begin{figure}[htbp]
    \centering
    \includegraphics[width=0.48\textwidth]{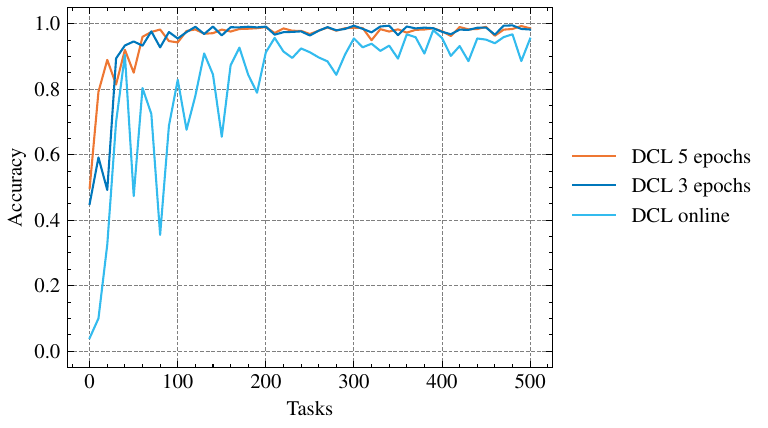} 
    \caption{Cumulative test accuracy of \acrlong{dcl} in the online scenario (every image is seen only once) and two offline scenarios (3 and 5 training epochs per task).}
    \label{fig:res:online_vs_offline}
\end{figure}

In all the main paper experiments, we applied our method in offline mode mode: we performed multiple training passes over the data for each task. However, efficiently learning from streaming data might require observing each training sample only once to make sure computation is not becoming a bottleneck. This is why we test our method in the online learning regime and compare it to two offline learning scenarios. The results are shown in \cref{fig:res:online_vs_offline}. Unsurprisingly, training for multiple epochs results in better and more robust accuracy on past tasks; it is however worth noting that our method still improves over time in the online learning scenario.

\end{document}